\documentclass[a4paper,twoside]{article}

\usepackage{epsfig}
\usepackage{subcaption}
\usepackage{calc}
\usepackage{amssymb}
\usepackage{amstext}
\usepackage{amsmath}
\usepackage{amsthm}
\usepackage{multicol}
\usepackage{pslatex}
\usepackage{tikz}
\usetikzlibrary{positioning}   
\usetikzlibrary{arrows.meta}   
\usetikzlibrary{calc}    
\usepackage{apalike}
\usepackage{algorithm2e}
\usepackage[bottom]{footmisc}
\usepackage{booktabs}
\usepackage{multirow}
\usepackage{dblfloatfix}
\usepackage{SCITEPRESS}
\usepackage{graphicx}
\usepackage{float}
\definecolor{myblue}{rgb}{0.21,0.49,0.74}\usepackage[colorlinks=true,allcolors=myblue]{hyperref}\usepackage[nameinlink,noabbrev]{cleveref}

\begin{document}

\title{Is Hierarchical Quantization Essential for Optimal Reconstruction?}

\author{Shirin Reyhanian\authorname{\sup{1}, Laurenz Wiskott\sup{1}}
\affiliation{\sup{1}Institute for Neural Computation (INI), Faculty of Computer Science, Ruhr University Bochum, Germany}
\email{shirin.reyhanian@ini.rub.de, laurenz.wiskott@rub.de}
}

\abstract{Vector-quantized variational autoencoders (VQ-VAEs) are central to models that rely on high reconstruction fidelity, from neural compression to generative pipelines. Hierarchical extensions, such as VQ‑VAE2, are often credited with superior reconstruction performance because they split global and local features across multiple latent levels. However, since higher-level latents derive all their information from lower levels, they should not carry additional reconstructive content beyond what the lower-level already encodes.  Combined with recent advances in training objectives and quantization mechanisms, this leads us to ask whether a single‑level VQ‑VAE, with matched representational budget and no codebook collapse, can equal the reconstruction fidelity of its hierarchical counterpart.  Although the multi-scale structure of hierarchical models may improve perceptual quality in downstream tasks, the effect of hierarchy on reconstruction accuracy, isolated from codebook utilization and overall representational capacity, remains empirically underexamined. We revisit this question by comparing a two‑level hierarchical VQ‑VAE and a capacity‑matched single‑level model on high‑resolution ImageNet images. Consistent with prior observations, we confirm that inadequate codebook utilization limits single‑level VQ‑VAEs and that overly high‑dimensional embeddings destabilize quantization and increase codebook collapse. We show that lightweight interventions such as initialization from data, periodic reset of inactive codebook vectors, and systematic tuning of codebook size and dimension significantly reduce collapse and enable the single-level model to make effective use of its available capacity. Our results demonstrate that when representational budgets are matched, and codebook collapse is mitigated, single‑level VQ‑VAEs can match the reconstruction fidelity of hierarchical variants, challenging the assumption that hierarchical quantization is inherently superior for high‑quality reconstructions. The code for reproducing our experiments is available at \url{https://github.com/wiskott-lab/single-vs-hier-recon}.}

\onecolumn \maketitle \normalsize \setcounter{footnote}{0} \vfill

\section{Introduction}
Reconstruction fidelity is a central objective in autoencoding systems with an information bottleneck, as it quantifies how much of the input signal is preserved after encoding.
In learned image compression, autoencoders are trained to balance bitrate and reconstruction error, so reconstruction quality measures the distortion incurred at a given bitrate budget~\cite{balle2017endtoend,balle2018hyperprior,minnen2018joint}. In latent generative modeling pipelines, the autoencoder acts as a tokenizer that maps images into a lower-dimensional (often discrete) representation on which a separate probabilistic model is trained, effectively decoupling representation learning from distribution modeling \cite{ramesh2021zero,esser2021taming,rombach2022ldm,fayyaz2022model}. In all such systems, assuming a sufficiently strong decoder, reconstruction error quantifies information discarded by the bottleneck and thus sets an upper bound on the achievable performance of any downstream module operating on the reconstruction or the latent representation.

Vector-Quantized Variational Autoencoders (VQ-VAEs) have become a popular approach for learning discrete latent representations by introducing a codebook-based quantization bottleneck into an autoencoder \cite{oord2017vqvae}. Continuous outputs of the encoder are quantized by mapping each feature vector to its nearest prototype in a learned codebook, yielding a latent map that can be represented as a grid of discrete indices. The representational capacity of this latent space is governed by the codebook capacity and the spatial resolution of the latent map, which together determine the maximum information content the bottleneck can convey. 

Optimizing VQ-VAEs is challenging due to the non-differentiable quantization. The original formulation uses the straight-through estimator and a commitment loss to enable end-to-end training~\cite{oord2017vqvae}, while EMA-based codebook updates have been shown to improve stability and code usage~\cite{razavi2019vqvae2,zeghidour2021soundstream}. Nonetheless, VQ-VAEs can suffer from codebook under-utilization (dead codebook vectors and low effective perplexity), which reduces effective capacity and can degrade reconstruction fidelity~\cite{takida2022sqvae,zheng2023online}. A growing body of work addresses this issue via improved objectives and codebook reset mechanisms designed to maintain codebook utilization~\cite{williams2020hqa,zeghidour2021soundstream,gu2022tokenizer,zheng2022movq,zheng2023online,takida2022sqvae}. 

Hierarchical variants of VQ-VAEs, first introduced in VQ-VAE-2 ~\cite{razavi2019vqvae2}, have been reported to improve reconstruction and generative quality by factorizing the representation across multiple quantized latent maps at different spatial scales. This design is commonly interpreted as a coarse-to-fine decomposition, where higher-level latents capture global layout and long-range structure, while lower-level latents refine local appearance and texture. However, the higher-level latents are derived from the lower-level representation and are therefore not an independent source of information. This motivates our hypothesis: Under matched representational budget and effective collapse mitigation, a single-level VQ-VAE should achieve reconstruction peformance comparable to a hierarchical counterpart. We note that hierarchical multi-scale structure may still offer other advantages, for instance for perceptual modeling or downstream generation, which we do not aim to rule out here.

In this work, we test this hypothesis via a controlled, capacity-matched comparison between a hierarchical VQ-VAE and a single-level VQ-VAE trained under an identical regime. We match the architectures in terms of both (1) the continuous latent map budget, defined by the number of latent vectors produced by the encoder and their embedding dimensionality, and (2) the discrete codebook budget, defined by the codebook size (number of entries) and codebook dimension (dimensionality of the codebook vectors). Both models are trained under an identical regime, using EMA-based vector quantization and the same procedures to mitigate codebook under-utilization. We summarize our contributions as follows:
\begin{itemize}
    \item To the best of our knowledge, we present the first controlled, capacity-matched comparison between single-level and hierarchical VQ-VAEs, designed to isolate the effect of hierarchical quantization on reconstruction fidelity.
    \item We show that a single-level VQ-VAE can achieve reconstruction fidelity comparable to a hierarchical VQ-VAE when the overall representational budget is matched and codebook collapse is mitigated using lightweight mechanisms.
    \item We quantitatively characterize codebook utilization in both architectures and identify codebook hyperparameter regimes under which single-level models exploit their codebook budget as effectively as hierarchical counterparts.
\end{itemize}

\section{Related Work}
\label{sec:related}

VQ-VAEs have introduced discrete autoencoding by quantizing encoder features via nearest-neighbor lookup in a learned codebook, representing images as spatial grids of discrete indices while enabling end-to-end learning through the straight-through estimator and auxiliary commitment and codebook losses~\cite{oord2017vqvae}. 

A persistent limitation of vector quantized models is codebook collapse, where only a subset of codebook entries are selected while many remain inactive. This reduces the effective discrete capacity and can limit reconstruction fidelity even when the nominal codebook capacity is large ~\cite{yu2021vector,zheng2022movq,esser2021taming,esser2021imagebart,chang2022maskgit,rombach2022ldm}. Beyond the original commitment regularization and EMA based codebook updates ~\cite{oord2017vqvae}, prior work has proposed several complementary strategies that explicitly improve codebook utilization.

Initialization and maintenance methods aim to keep codebook entries aligned with the encoder output distribution. A common strategy is to monitor code usage during training and periodically refresh inactive entries by reinitializing them with recent encoder outputs or feature statistics. Zheng and Vedaldi have proposed online codebook learning that reset dead codes using encoded features~\cite{zheng2023online}. Similarly, SoundStream uses k-means on early encoder outputs for initialization, and periodically replaces inactive codebook vectors~\cite{zeghidour2021soundstream}.

Design choices for the assignment space aim to promote broader codebook utilization by shaping the quantization space, such as by projecting to a lower-dimensional quantization space and using cosine-similarity matching~\cite{yu2022vqgan,zhu2024scaling}. Related work also replaces learned codebook lookup with fixed structured quantizers. Finite Scalar Quantization discretizes each projected latent dimension to a small set of fixed levels, yielding an implicit Cartesian grid of codes in a bounded hypercube and reducing reliance on codebook updates~\cite{mentzer2023finite}. Lookup Free Quantization similarly maps latents directly to fixed binary levels, producing codes that lie on the vertices of a hypercube~\cite{yu2023language,yu2023magvit}.

Optimization and estimator improvements address collapse at its source. Huh et al.\ have analyzed pathologies of the straight-through estimator and have proposed modified parameterizations~\cite{huh2023straightening}. Takida et al.\ study stochastic quantization to stabilize learning and reduce collapse~\cite{takida2022sqvae}. SimVQ introduces a simple linear reparameterization to mitigate collapse by optimizing a shared latent space~\cite{zhu2025addressing}, and the rotation trick modifies gradient flow through quantization and reports consistent gains in utilization and reconstruction metrics~\cite{fifty2025restructuring}. 


Hierarchical VQ-VAE models distribute discrete capacity across multiple quantization stages, for example through multi-scale latent maps at different spatial resolutions or through stacked residual quantizers, to preserve global structure and fine detail under a compact discrete budget. VQ-VAE2 introduced a multi-level hierarchy that combines coarse and fine discrete latents and reported improved reconstruction and generation on complex image datasets such as ImageNet \cite{razavi2019vqvae2}. Related hierarchical formulations have been studied in extreme compression regimes to retain perceptual quality and semantic content at very low bitrates \cite{williams2020hqa,NEURIPS2024_9a24e284}. Beyond images, hierarchical models have been used to compress long-context signals prior to sequence modeling, such as Jukebox for music generation and VideoGPT for video generation ~\cite{dhariwal2020jukebox,yan2021videogpt}.

These hierarchical constructions introduce additional design degrees of freedom, like how capacity is split across levels, how dependencies between levels are parameterized, and how multiple codebooks are optimized and utilized, and can suffer from level-wise codebook collapse, where some levels contribute little despite added complexity~\cite{williams2020hqa}. Prior studies report that effectively routing information to higher latent levels in hierarchical models is challenging and often requires additional architectural or training components \cite{dhariwal2020jukebox}. HQ-VAE targets these training challenges and frames hierarchical discrete learning in a variational Bayes formulation and reports improved code usage and reconstruction in hierarchical settings~\cite{takida2023hqvae}.

Although hierarchical models are often assumed to yield superior reconstructions, the empirical evidence is frequently confounded by differences in total representational budget, encoder-decoder strength, loss functions, and codebook utilization. As a result, it remains unclear to what extent reconstruction gains should be attributed to hierarchical factorization per se versus increased effective capacity and improved utilization under different training recipes.

Our study directly targets this gap through a controlled, capacity-matched comparison focused on reconstruction fidelity. Concretely, we identify codebook hyperparameter regimes under which a single-level VQ-VAE exploits its discrete budget as effectively as a hierarchical counterpart. Our results suggest that three lightweight interventions, namely initialization from data, periodic reset of inactive codes based on statistics aggregated over recent batches, and systematic tuning of codebook size and code-vector dimensionality, enable single-level VQVAE to match hierarchical reconstruction accuracy under matched budgets. These interventions are simple to implement and can be integrated into existing training pipelines with minimal changes.
\section{Methodology}
\label{sec:method}
\subsection{Vector Quantized Autoencoding}

A VQ-VAE consists of an encoder $E_{\phi}$, a vector quantizer with a discrete codebook, and a decoder $G_{\theta}$. Given an input image $x$, a continuous latent representation $z_o = E_{\phi}(x) \in \mathbb{R}^{H_e \times W_e \times C_e}$ is produced by the encoder.
The latent representation is then projected into the quantization space using a learned $1 \times 1$ convolution $P_{\omega}$:
\begin{equation}
z_e = P_{\omega}(z_o) \in \mathbb{R}^{H_e \times W_e \times D}
\end{equation}

For each spatial location $(i,j)$ in the latent representation, the corresponding latent vector is denoted by $\mathbf{z}_{e,ij} \in \mathbb{R}^{D}$. Each latent vector is then assigned to its nearest codebook entry:
\begin{equation}
k(i,j) = \arg\min_{k \in \{1,\dots,K\}} \lVert \mathbf{z}_{e,ij} - \mathbf{c}_k \rVert_2^2,
\qquad
\mathbf{z}_{q,ij} = \mathbf{c}_{k(i,j)}
\end{equation}
Collecting all locations yields the quantized latent representation $z_q \in \mathbb{R}^{H_e \times W_e \times D}$, which is decoded to obtain the reconstruction $\hat{x} = G_{\theta}(z_q).$ 

Quantization is non-differentiable, therefore, similar to VQ-VAE2~\cite{razavi2019vqvae2} we combine the straight-through estimator with EMA codebook updates. In the forward pass, the decoder processes the quantized representation $z_q$ to reconstruct the input image. In the backward pass, gradients are passed to the encoder through $z_e$ using the straight-through estimator, treating the quantization step as the identity for gradient propagation. All models in this study are trained with the same objective and identical update hyperparameters. Concretely, we minimize
\begin{equation}
\mathcal{L}
=
\lVert x - \hat{x} \rVert_2^2
+
\beta \lVert z_e - \mathrm{sg}(z_q) \rVert_2^2
\end{equation}
where $\mathrm{sg}(\cdot)$ denotes the stop gradient operator and $\beta$ is the commitment weight.

\paragraph{Single-level VQ-VAE}
The single level model uses a single latent representation $z_{o,s} = E_{\phi_s}(x)\in \mathbb{R}^{H_s \times W_s \times C_s}$ with a codebook of size $K_s$ and dimension $D_s$. After projection and quantization, the $z_{q,s} \in \mathbb{R}^{H_s \times W_s \times D_s}$ is used to reconstructs $\hat{x} = G_{\theta_s}(z_{q,s})$.

\paragraph{Two-level hierarchical VQ-VAE}
The hierarchical model uses a two-level factorization with a bottom latent representation $z_{o,b} = E_{\phi_b}(x) \in \mathbb{R}^{H_b \times W_b \times C_h}$ and a top latent representation $z_{o,t} = E_{\phi_t}(z_{o,b}) \in \mathbb{R}^{H_t \times W_t \times C_h}$ at a coarser spatial scale, where $C_h$ is equal for both levels. Each level is quantized with its own codebook, of the same size $K_h$, and the same dimension $D_h$. The resulting quantized latent representations are combined as $z_{q,h} = \operatorname{concat}\big(z_{q,b}, \operatorname{upsample}(z_{q,t})\big)$ and decoded to reconstruct $\hat{x} = G_{\theta_h}(z_{q,h})$.

Razavi et al.~\cite{razavi2019vqvae2} further propose conditioning the lower-level representation on the upsampled higher-level representation using additional decoder modules, with the goal of reducing redundancy and promoting a coarse to fine decomposition \cite{razavi2019vqvae2}. Prior work reports that this conditioning can shift reconstruction relevant information from the lower level to the higher level \cite{reyhanian2024analysis}. In our experiments, we evaluated conditioned and unconditioned hierarchical variants and observed no measurable difference in overall reconstruction accuracy. This indicates that, for reconstruction accuracy, the decoder primarily benefits from the aggregate information provided by the discrete latents rather than from how that information is distributed across levels. We therefore report results for the unconditioned hierarchical model in the remainder of the paper for simplicity.

\subsection{Representational capacity matching}Our goal is to isolate the effect of hierarchical quantization on reconstruction accuracy. To avoid confounding from unequal representational capacity, we match both architectures along two factors, namely continous latent budget and discrete codebook budget. 

\paragraph{Continuous latent budget} We define the continuous latent budget for each level as $H\times W\times C$. With $H_s W_s = H_b W_b$, we choose $C_s$ and $C_h$ such that:
\begin{equation}
H_s W_s C_s = H_b W_b C_h + H_t W_t C_h
\label{eq:4}
\end{equation}
So the single-level and hierarchical models have matched continuous capacity at the quantizer input.

\paragraph{Discrete codebook budget} We define the discrete codebook budget of a quantizer as the product of codebook size and code vector dimensionality, $K D$. In the hierarchical model, both levels use the same size and dimension, so the total discrete budget is $2K_hD_h$. We match the single-level codebook budget to the total discrete budget of hierarchical model as follows:
\begin{equation}
K_s D_s = 2 K_h D_h
\label{eq:5}
\end{equation}

\subsection{Codebook collapse mitigation}

Capacity matching is insufficient when the model suffers from codebook collapse and does not use its available discrete capacity. Consistent with prior observations, we also observe that inactive codebook vectors and low perplexity can limit reconstruction quality. To ensure a fair comparison, we apply lightweight procedures that mitigate codebook collapse:
\begin{itemize}
    \item \textbf{Initialization from input} Codebook vectors are initialized from encoder outputs on randomly selected training samples, so optimization starts from representative prototypes.
    \item \textbf{Periodic dead-code reset} Code vectors that receive very few assignments over a sliding window of batches are reset using recently observed encoder outputs.
    \item \textbf{Codebook hyperparameter tuning} We conduct targeted experiments that vary codebook size and dimension to assess their impact on quantization stability and utilization. Based on these experiments, we provide practical tuning guidelines that avoid collapse prone regimes and identify settings where the single-level VQ-VAE matches the reconstruction accuracy of its hierarchical counterpart.
    
\end{itemize}
These procedures leave the reconstruction objective unchanged and maintain effective discrete capacity, so reconstruction differences can be attributed to architectural hierarchy rather than codebook under-utilization. Moreover, they are lightweight mechanisms that require only minor changes to existing training pipelines.

\section{Experiments and Results}
\label{sec:results}

This section evaluates reconstruction fidelity of single-level and hierarchical VQ-VAE models under controlled, capacity-matched conditions.

\paragraph{Training and evaluation setup}
All models are trained and evaluated on ImageNet with images resized to $256 \times 256$. Training is run for 300 epochs on 4 NVIDIA A100 GPUs (40\,GB). Optimization uses Adam with learning rate $3\times10^{-4}$ and exponential decay factor $0.99$.

\paragraph{Model configurations}
The hierarchical model uses a two-level latent representation with resolutions $H_b \times W_b = 64 \times 64$, $H_t \times W_t = 32 \times 32$, and $C_h = 128$ at both levels. The single-level model uses a single latent representation with resolution $H_s \times W_s = 64 \times 64$ and $C_s = 160$. For all experiments, this configuration for continuous latent budget is fixed and matched between architectures according to \Cref{eq:4}. The discrete codebook budget is matched as in \Cref{eq:5}; within this constraint, codebook size and dimension are varied to identify regimes that improve codebook utilization and mitigate collapse. All remaining details, including the collapse mitigation procedures, follow \Cref{sec:method}. Both architectures are trained with the same reconstruction objective, EMA quantizer, and optimization schedule, so performance differences can be attributed to the presence or absence of hierarchical quantization.

\begin{figure}[htp]
    \centering
    \includegraphics[width=.85\columnwidth]{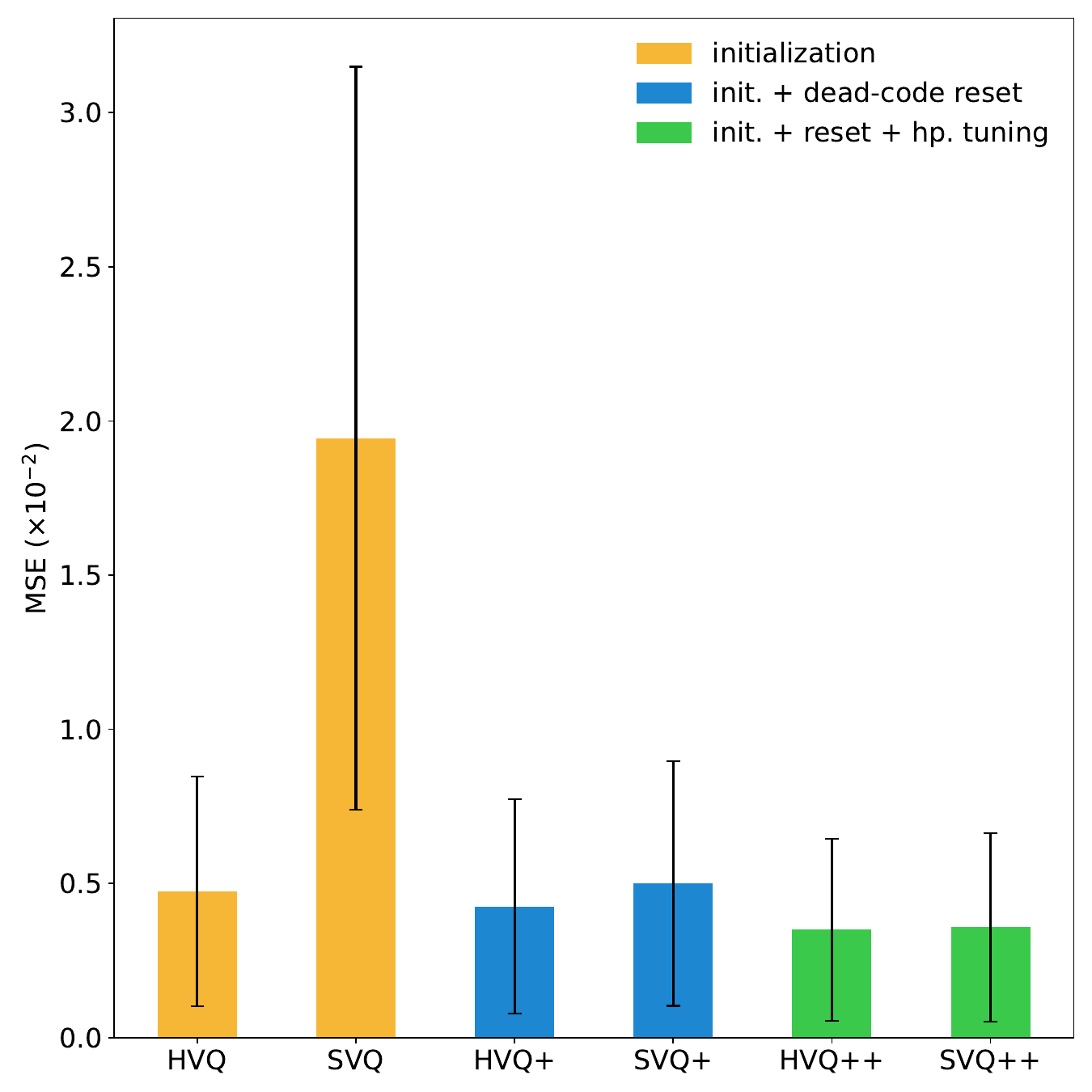}
    \includegraphics[width=.845\columnwidth]{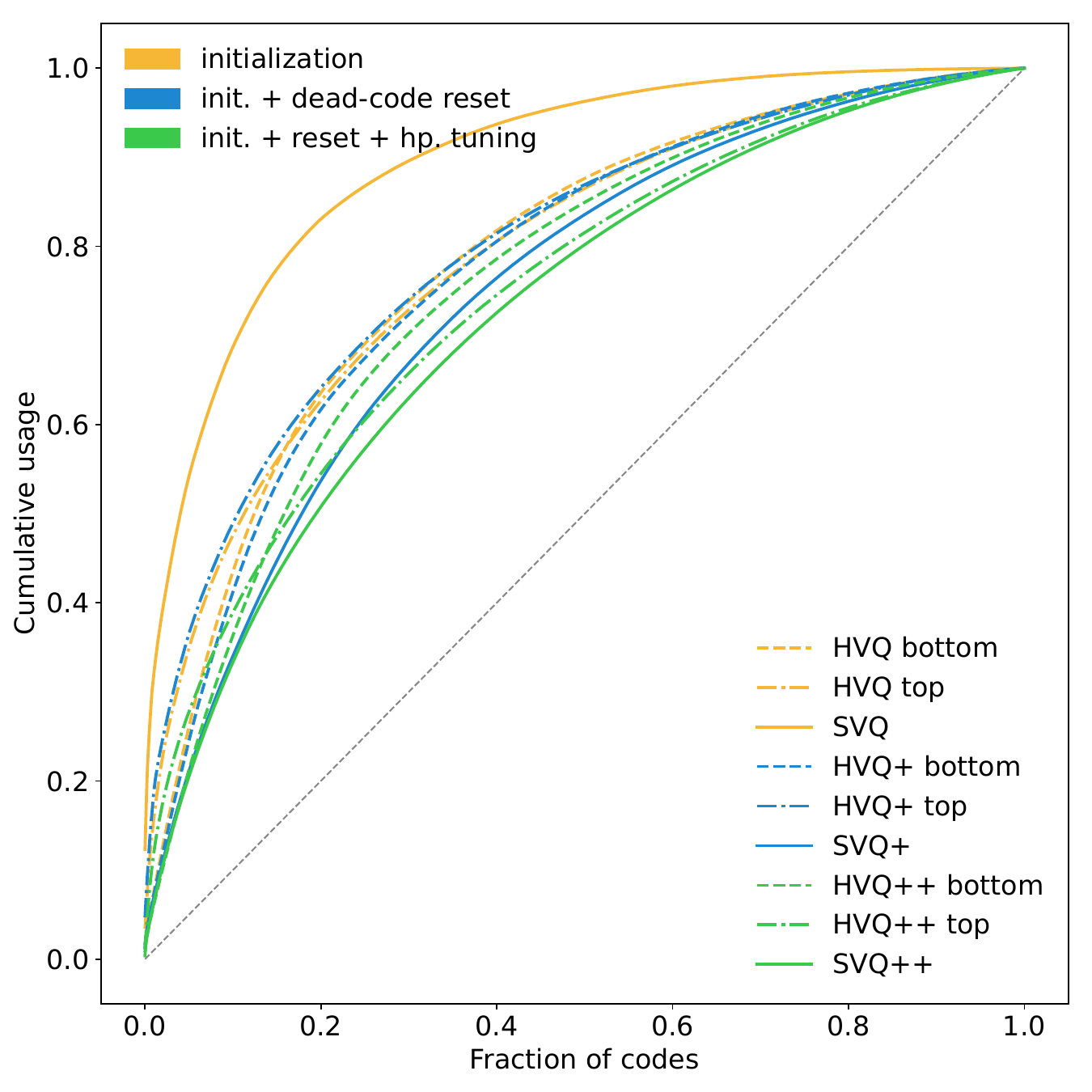}
    \includegraphics[width=.85\columnwidth]{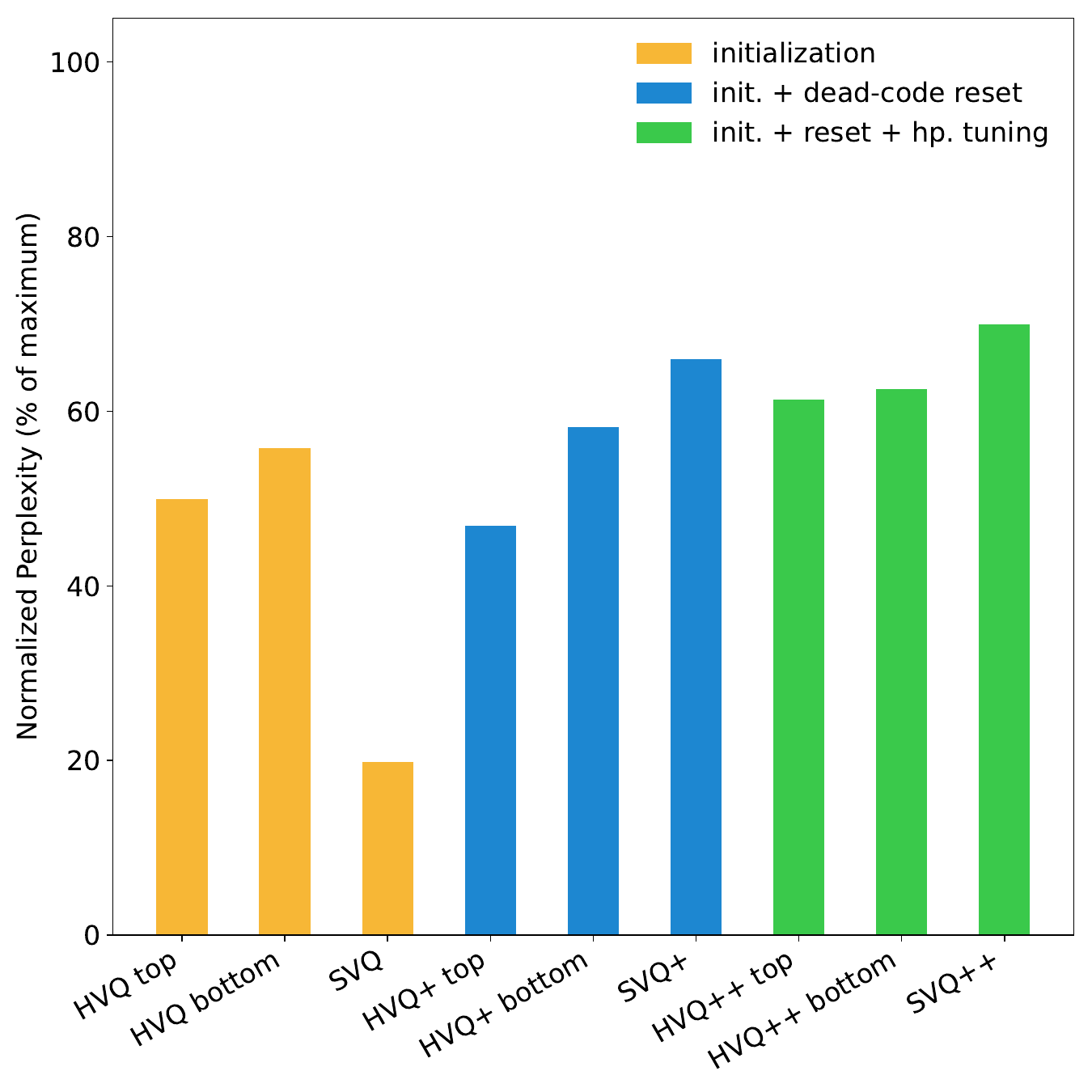}
    \caption{Ablation of codebook collapse interventions. \textbf{Top:} Average reconstruction error, \textbf{Middle: }Lorenz curves, and \textbf{Bottom:} Normalized perplexity for single-level and hierarchical VQ-VAE models with matched representational budgets on the ImageNet validation set.}
    \label{fig:three_stack}
\end{figure}

\begin{figure*}[htp]
    \centering
    \includegraphics[width=\textwidth]{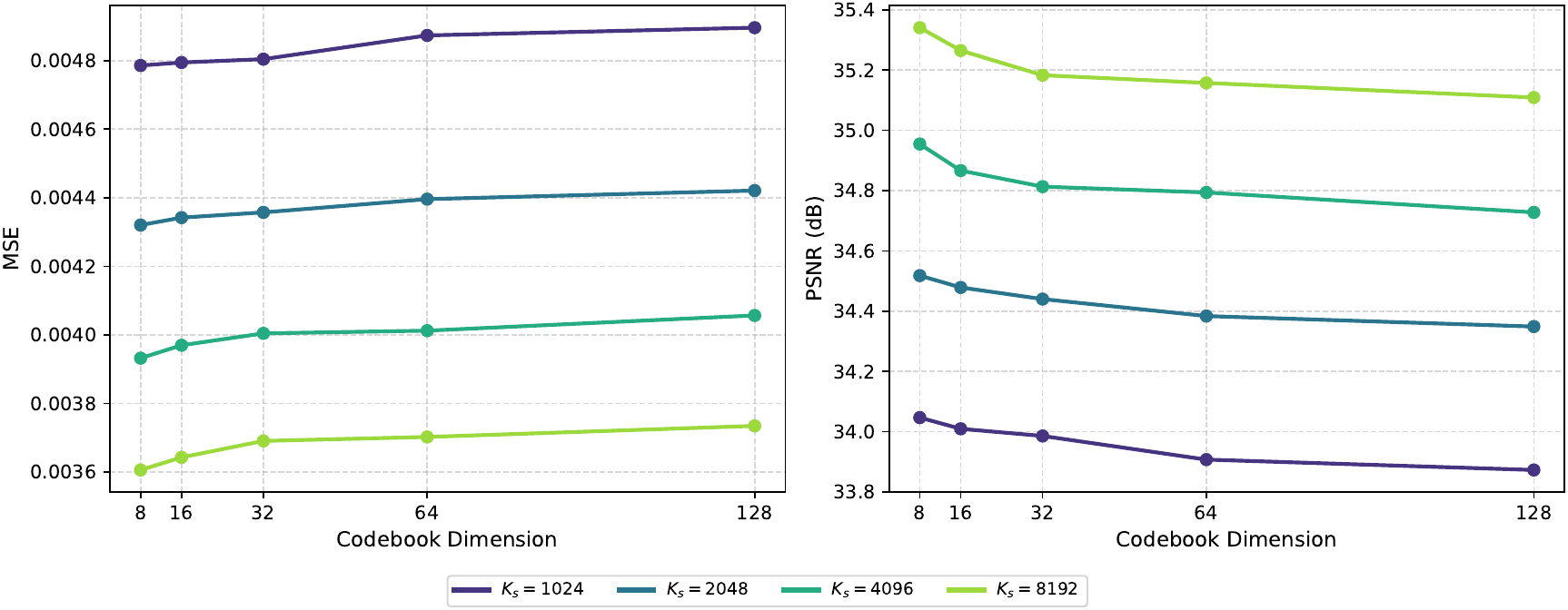}
    \caption{Effect of codebook size and dimension on reconstruction accuracy in single-level VQ-VAE.}
    \label{fig:kd_sweep}
\end{figure*}

\subsection{Mitigating codebook collapse}
\label{sec:results_collapse}

Capacity matching is only meaningful when the codebook is actually used. Consistent with prior observations, we find that inactive codebook vectors and low perplexity correlate with degraded reconstruction quality. To reduce this confounder, three lightweight interventions are applied throughout. Firstly, code vectors are initialized from encoder outputs on random training samples. Secondly, inactive codebook vectors are detected when their assignment count stays below a threshold over a recent window of batches. When this happens, the code vector is reset using an average from recently observed encoder outputs. We sweep thresholds from 1 to 5 and find 2 to work best in our setting. The window size is fixed to 10 batches. Thirdly, we also tune codebook hyperparameters by reducing the dimensionality of the quantization space in a representative capacity-matched single-level and hierarchical pair, to assess how each intervention affects reconstruction accuracy.~\Cref{fig:three_stack} reports an ablation of the proposed interventions. Periodic code resets improve reconstruction for both architectures, indicating that codebook collapse affects both. Reducing dimensionality from $64$ to $16$, while keeping the codebook budget constant by increasing the codebook size according to \Cref{eq:5}, yields an additional gain for both models. 

To connect reconstruction fidelity to discrete capacity usage, we report average MSE, perplexity and Lorenz curves for the representative models in~\Cref{fig:three_stack}. Perplexity is defined as the exponential of the Shannon entropy of the empirical code-assignment distribution and can be interpreted as the effective number of active codebook vectors. The Lorenz curve is obtained by sorting codebook vectors by assignment frequency and plotting the cumulative share of assignments against the cumulative fraction of codes, where curves closer to the diagonal indicate more uniform code usage.

As shown in~\Cref{fig:three_stack}, each intervention reduces MSE and is accompanied by higher perplexity and a less concentrated assignment distribution, indicating that improved codebook utilization is a key driver of reconstruction fidelity in both architectures. With all interventions enabled, the remaining gap between single-level and hierarchical reconstruction becomes minimal, motivating a broader sweep over codebook size and dimension.

\subsection{Effect of codebook size and dimension}
\label{sec:results_hyperparams}

We have performed a grid sweep over codebook size $K_s \in \{1024, 2048, 4096, 8192\}$ and code vector dimension  $D_s \in \{8,16,32,64,128\}$ for the single-level VQ-VAE, and report PSNR and MSE averaged over the ImageNet validation set. \Cref{fig:kd_sweep} summarizes the results. Increasing $K_s$ yields a consistent improvement in reconstruction accuracy. In contrast, increasing $D_s$ generally degrades performance, even though it increases the nominal discrete capacity. The best results are obtained at $D_s=8$, while $D_s=128$ performs worst. 

This behavior is consistent with the interpretation that a lower-dimensional quantization space encourages broader and more stable code usage, whereas overly high-dimensional code vectors exacerbate under-utilization and hinder effective assignments. We observed that reducing $D_s$ below $8$ leads to a sharp drop in reconstruction quality; these settings are omitted from \Cref{fig:kd_sweep} to preserve readability. For reference, the hierarchical model evaluated at representative settings, including $D_h=8$ and $D_h=128$, follows the same trend.

These results suggest a practical tuning rule for reconstruction-focused models. When additional discrete capacity is needed, increasing the codebook size is generally preferable to increasing the code vector dimensionality.

\begin{table*}[t]
\centering
\scriptsize
\setlength{\tabcolsep}{4pt}
\renewcommand{\arraystretch}{1.08}

\caption{Quantitative comparison of single-level and hierarchical VQ-VAE reconstructions under matched representational budgets. 
Metrics are reported as mean $\pm$ std on the ImageNet validation set. 
Params denotes the number of trainable parameters.}
\label{tab:capacity_matched}

\begin{tabular*}{\textwidth}{@{\extracolsep{\fill}} c c c c c c c c c}
\toprule
\multicolumn{3}{c}{Codebook parameters} & \multicolumn{3}{c}{Single-level} & \multicolumn{3}{c}{Hierarchical} \\
\cmidrule(lr){1-3}\cmidrule(lr){4-6}\cmidrule(lr){7-9}
$D_s=D_h$ & $K_s$ & $2\times K_h$ &
PSNR $\uparrow$ & MSE $\downarrow$ & \#Params (M) &
PSNR $\uparrow$ & MSE $\downarrow$ & \#Params (M) \\
\midrule
8   & 1024 & $2\times512$  & 34.1 $\pm$ 3.5 & 0.004 $\pm$ 0.004 & 1.1 & 34.5 $\pm$ 3.5 & 0.004 $\pm$ 0.004 & 1.1 \\
8   & 2048 & $2\times1024$ & 34.5 $\pm$ 3.5 & 0.004 $\pm$ 0.003 & 1.1 & 34.9 $\pm$ 3.6 & 0.004 $\pm$ 0.003 & 1.1 \\
8   & 4096 & $2\times2048$ & 35.0 $\pm$ 3.6 & 0.003 $\pm$ 0.003 & 1.1 & 35.4 $\pm$ 3.6 & 0.003 $\pm$ 0.003 & 1.1 \\
8   & 8192 & $2\times4096$ & 35.3 $\pm$ 3.6 & 0.003 $\pm$ 0.003 & 1.1 & 35.8 $\pm$ 3.6 & 0.003 $\pm$ 0.003 & 1.1 \\
\midrule
16  & 8192 & $2\times4096$ & 35.3 $\pm$ 3.5 & 0.003 $\pm$ 0.003 & 1.1 & 35.7 $\pm$ 3.6 & 0.003 $\pm$ 0.003 & 1.1 \\
32  & 8192 & $2\times4096$ & 35.2 $\pm$ 3.5 & 0.003 $\pm$ 0.003 & 1.2 & 35.6 $\pm$ 3.6 & 0.003 $\pm$ 0.003 & 1.2 \\
64  & 8192 & $2\times4096$ & 35.2 $\pm$ 3.5 & 0.003 $\pm$ 0.003 & 1.2 & 35.6 $\pm$ 3.5 & 0.003 $\pm$ 0.003 & 1.3 \\
128 & 8192 & $2\times4096$ & 35.1 $\pm$ 3.3 & 0.003 $\pm$ 0.004 & 1.4 & 35.4 $\pm$ 3.6 & 0.003 $\pm$ 0.003 & 1.7 \\
\bottomrule
\end{tabular*}
\end{table*}

\subsection{Capacity matched single-level vs. hierarchical models}
\label{sec:results_matched}
\begin{figure}[b]
    \centering
    \includegraphics[width=\columnwidth,keepaspectratio]{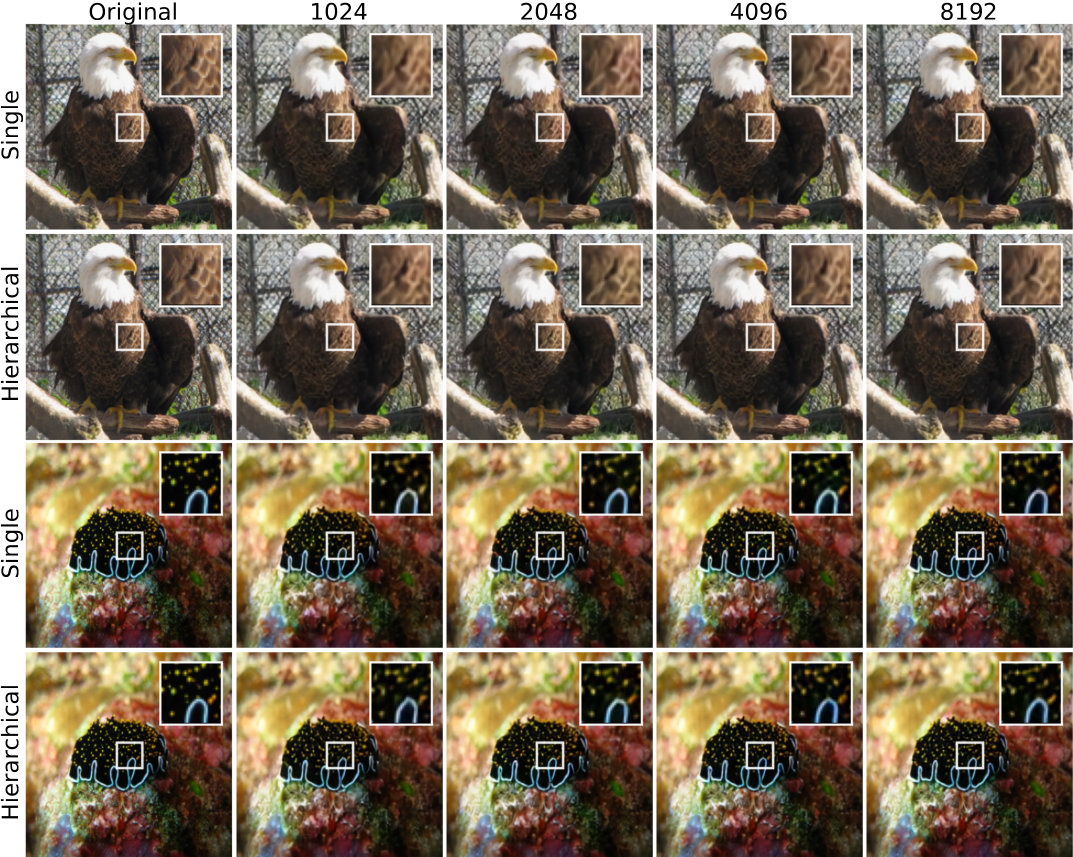}
    \caption{ Qualitative reconstructions across single-level and hierarchical VQ-VAE models. The first column shows the original images. Columns 2-5 depict reconstructions obtained with varying codebook sizes $K_s=2K_h$. All models use Codebook dimension $D_s=D_h=8$.}
    \label{fig:2}
\end{figure}

After identifying codebook hyperparameter regimes that perform well for both architectures, we compare single-level and hierarchical VQ-VAEs under matched continuous and discrete budgets in two controlled studies. First, we fix the code vector dimensionality to $D_s = D_h = 8$ and sweep the codebook size. Second, motivated by the consistent gains from larger codebooks, we fix the codebook size to $K_s = 8192$ for the single-level model and $2K_h = 8192$ for the hierarchical model, and sweep the code vector dimensionality for both. We summarize the quantitative results for each architecture under matched budgets in \Cref{tab:capacity_matched} and present a qualitative comparison in \Cref{fig:2}.

Across these matched-budget comparisons, both pixel-level metrics and qualitative reconstructions indicate that the single-level VQ-VAE achieves reconstruction performance comparable to its hierarchical counterpart when trained in the utilization-stable regimes identified above. This supports our central hypothesis: Since the top-level representations are computed from representations derived from the bottom level, hierarchical factorization does not provide an additional independent source of reconstructive information. Instead, the reconstruction advantage often attributed to hierarchy can be explained by differences in how effectively the available representational capacity is utilized. Once codebook under-utilization is mitigated and representational budgets are matched, both architectures can encode the same reconstructive content with similar fidelity.

Overall, these results suggest that hierarchy is not inherently required for optimal reconstruction accuracy in VQ-VAE models. For applications that prioritize architectural simplicity and ease of deployment, a single-level model trained with lightweight utilization mechanisms provides a competitive alternative to hierarchical designs under equivalent capacity.
\section{Discussion}
\label{sec:discussion}

We have studied the role of hierarchical quantization in VQ-VAE models under a controlled, capacity-matched protocol. By matching continuous and discrete budgets and holding the training objective, EMA quantizer, and codebook collapse interventions fixed across architectures, we isolate the architectural effect of hierarchy on reconstruction. In utilization-stable regimes, the single-level VQ-VAE achieves reconstruction accuracy comparable to the hierarchical model on pixel-level metrics, suggesting that hierarchy is not necessary for optimal reconstruction under matched capacity.

Our hyperparameter sweep provides a practical parameter tuning guideline. Increasing codebook size improves reconstruction consistently, whereas increasing code vector dimensionality often degrades performance despite increasing nominal capacity. This indicates that, for reconstruction-focused models, expanding the codebook size is a more effective way to scale capacity than increasing codebook dimensionality.

These results favor single-level models in applications where the discrete bottleneck is used primarily for reconstruction, such as learned compression and representation learning components, since they offer comparable fidelity with fewer architectural choices and design complexities. 

Future work should conduct similar capacity-matched investigations to assess how hierarchical quantization affects autoregressive and diffusion priors trained on quantized representations, compared to their single-level counterparts.

\section*{\uppercase{Acknowledgements}}

This work was funded by the Deutsche Forschungsgemeinschaft (DFG, German Research Foundation) with funding for the research group FOR 2812 \textit{Constructing scenarios of the past: A new framework in episodic memory} by the grant 397530566 (P5).

We also acknowledge the Gauss Centre for Supercomputing e.V. (www.gauss-centre.eu) for funding this project by providing computing time through the John von Neumann Institute for Computing (NIC) on the GCS Supercomputer JUWELS~\cite{JUWELS} at Jülich Supercomputing Centre (JSC).

\bibliographystyle{apalike}
{\small
\bibliography{main}}


\end{document}